\documentclass[letterpaper, 10 pt, conference]{ieeeconf}
\IEEEoverridecommandlockouts
\usepackage{graphicx} 
\usepackage{ragged2e} 
\usepackage{array}
\usepackage{multirow}
\usepackage{booktabs} 
\usepackage{url}
\usepackage{xcolor}
\usepackage{pifont}
\usepackage{dirtree}

\usepackage[nocompress]{cite}

\usepackage[hidelinks]{hyperref}
\hypersetup{
    colorlinks=true,
    linkcolor=blue,
    citecolor=blue,
    urlcolor=blue,
    pdfstartview=FitH,
    pdfauthor={Your Name},
}

\usepackage{fancyhdr}
\fancypagestyle{withfooter}{
  
  \fancyhead[L]{}
  \fancyhead[R]{}
  \fancyfoot[C]{\footnotesize Presented at the 2025 IEEE ICRA Workshop on Field Robotics}
}

\newcommand{\cmark}{\textcolor[rgb]{0,0.5,0}{\ding{51}}} 
\newcommand{\xmark}{\textcolor[rgb]{0.6,0,0}{\ding{55}}} 

\title{\LARGE \bf
     CU-Multi: A Dataset for Multi-Robot Data Association
}

\author{Doncey Albin$^{1}$, Miles Mena$^{1}$,  Annika Thomas$^{2}$, Harel Biggie$^{3}$, Xuefei Sun$^{1}$, Dusty Woods$^{1}$ \\ Steve McGuire$^{4}$, and Christoffer Heckman$^{1}$
\thanks{$^{1}$Authors are with the Autonomous Robotics and Perception Group in the Computer Science Department at the University of Colorado Boulder, Boulder, CO 80309, USA}%
\thanks{$^{2}$Annika Thomas with the Aerospace Controls Laboratory at Massachusetts Institute of Technology, Cambridge, MA, 02139, USA}%
\thanks{$^{3}$Harel Biggie with the Computer Science and Artificial Intelligence Laboratory at Massachusetts Institute of Technology, Cambridge, MA 02139, USA}%
\thanks{$^{4}$Steve McGuire with the Human-Aware Robotic Exploration Lab at University of California Santa Cruz, Santa Cruz, CA 95064, USA}}
\begin{document}

\maketitle
\thispagestyle{withfooter}
\pagestyle{withfooter}


\begin{abstract}
Multi-robot systems (MRSs) are valuable for tasks such as search and rescue due to their ability to coordinate over shared observations. A central challenge in these systems is aligning independently collected perception data across space and time-- i.e., multi-robot data association. While recent advances in collaborative SLAM (C-SLAM), map merging, and inter-robot loop closure detection have significantly progressed the field, evaluation strategies still predominantly rely on splitting a single trajectory from single-robot SLAM datasets into multiple segments to simulate multiple robots. Without careful consideration to how a single trajectory is split, this approach will fail to capture realistic pose-dependent variation in observations of a scene inherent to multi-robot systems. To address this gap, we present CU-Multi, a multi-robot dataset collected over multiple days at two locations on the University of Colorado Boulder campus. Using a single robotic platform, we generate four synchronized runs with aligned start times and deliberate percentages of trajectory overlap. CU-Multi includes RGB-D, GPS with accurate geospatial heading, and semantically annotated LiDAR data. By introducing controlled variations in trajectory overlap and dense lidar annotations, CU-Multi offers a compelling alternative for evaluating methods in multi-robot data association. Instructions on accessing the dataset, support code, and the latest updates are publicly available at \url{https://arpg.github.io/cumulti}.
\end{abstract}

\section{INTRODUCTION}
\begin{figure}[thpb]
  \centering
  \includegraphics[width=1.0\linewidth]{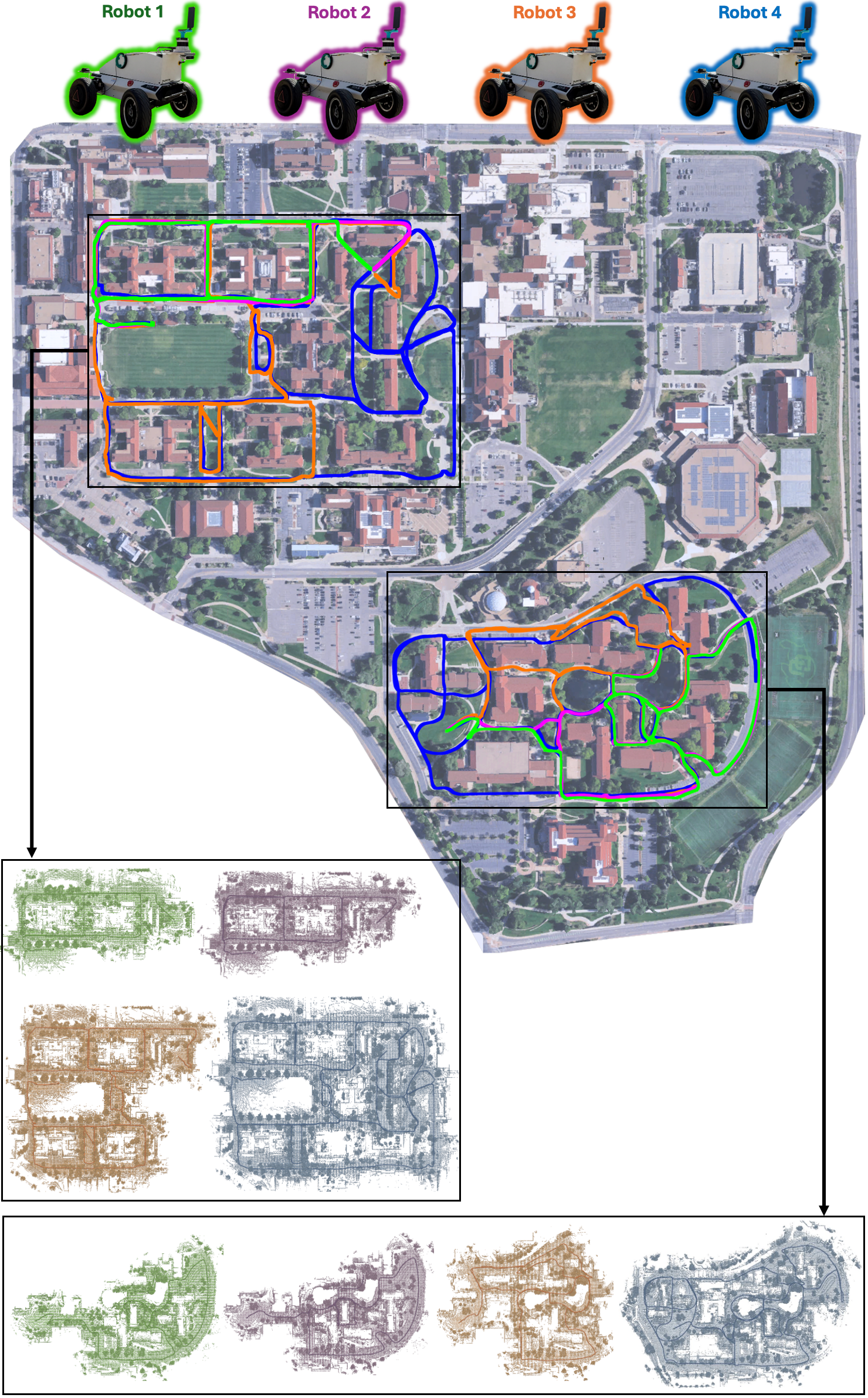}
  \caption{Overhead view of the all paths overlaid on map from the Main Campus (top) and Kittredge Loop (bottom) environments in the CU-Multi Dataset.}
  \label{fig:main}
\end{figure}
\begin{table*}
    \caption{Recent multi-robot data association methods and the types of data used for validation.}
    \label{tab:cslam_verification}
    \centering
    \begin{tabular}{ 
        >{\RaggedRight}m{2.5cm} 
        >{\centering\arraybackslash}m{4cm} 
        >{\centering\arraybackslash}m{3cm}
        >{\centering\arraybackslash}m{2.5cm}
        >{\centering\arraybackslash}m{2cm}
    } 
        \toprule
        \textbf{Method} & \textbf{Single-Robot Datasets Used with Trajectory Splitting} & \textbf{Multi-Robot Datasets Used} & \textbf{Multi-Robot Field Experiments} & \textbf{Simulation Experiments} \\[0.5ex]
        \midrule
        DOOR-SLAM \cite{lajoie2020door} & KITTI \cite{geiger2013vision} (seq 00) & \xmark & \cmark & \cmark \\[1ex]
        SWARM-SLAM \cite{lajoie2023swarm} & KITTI \cite{geiger2013vision} (seq 00), KITTI-360 \cite{liao2022kitti} (seq 09) & GRACO \cite{zhu2023graco}, S3E \cite{feng2024s3e} & \cmark & \xmark \\[1ex]
        CURB-SG \cite{greve2024collaborative} & \xmark & \xmark & \xmark & \cmark \\[1ex]
        DiSCo-SLAM \cite{huang2021disco} & KITTI \cite{geiger2013vision} (seq 00 and 08), Stevens Dataset \cite{legoloam2018} & Park Dataset & \xmark & \xmark \\[1ex]
        Multi S-Graphs \cite{fernandez2024multi} & \xmark & \xmark & \cmark & \cmark \\[1ex]
        Kimera-Multi \cite{tian2022kimera}  & \xmark & EuRoC Dataset \cite{burri2016euroc} & \cmark & \cmark \\[1ex]
        DCL-SLAM \cite{zhong2023dcl}  & KITTI \cite{geiger2013vision} (seq 05, 08, 09) & \xmark & \cmark & \xmark \\[1ex]
        \bottomrule
    \end{tabular}
\end{table*}

Multi-robot systems (MRSs) significantly enhance capabilities across diverse domains, particularly in large-scale environments, by accelerating exploration through distributed sensing and collaborative decision-making \cite{kim2023multi, marvel2018multi}. A central challenge in realizing these advantages lies in fusing perception data collected independently by multiple robots into a unified global representation, complicated by spatial and temporal misalignment. This challenge is further compounded in many practical scenarios where external positioning methods (e.g. GPS, motion capture) are impractical, unreliable, or hazardous. Multi-robot data association algorithms are typically first developed and validated offline using datasets before doing field tests in real-world conditions, where system-level issues like communication restrictions may arise. Consequently, the availability of realistic, well-structured multi-robot datasets is essential for supporting reproducible research and bridging the gap between offline development and real-world deployment.

Multi-robot data association underpins key methods such as global localization \cite{ankenbauer2023global, cho2022openstreetmap, thomas2024sosmatch}, collaborative SLAM (C-SLAM) \cite{zhao2023review}, map merging \cite{yu2020review, yin2023automerge, stathoulopoulos2024frame}, and inter-robot loop-closure detection \cite{wang2023wi}. Despite recent advancements in these methods, there remains a prevalent practice of artificially segmenting a single trajectory into multiple parts to simulate a multi-robot scenario for verification \cite{zhong2023dcl} (Table \ref{tab:cslam_verification}). Besides the availability of both camera and lidar semantics, the trajectories from single-robot datasets can be quite large, including a variety of loop-closure types desired for testing \cite{zhong2023dcl, lajoie2023swarm}. However, without careful consideration, arbitrary segmentation can fail to accurately represent realistic observational overlap typically encountered in multi-robot operations \cite{lajoie2022towards}. Furthermore, as we discuss in the following section, no standardized method currently exists for such trajectory partitioning. 

Several multi-robot datasets have recently been introduced for C-SLAM verification \cite{feng2024s3e, zhou2024coped, zhu2023graco}. These datasets include temporally synchronized multi-agent sensor data collected across both indoor and/or outdoor environment, as well as support varying degrees of inter-robot interaction and loop closure scenarios. While these datasets represent strong starting points, there remains a need for multi-robot datasets that incorporate dense point-wise semantic annotations for LiDAR and explicitly varied trajectory overlap (see Table~\ref{tab:multi_robot_datasets}). Trajectory overlap is a key factor in selecting appropriate data association strategies \cite{lajoie2023swarm}, yet few datasets explicitly account for this. Moreover, despite the availability of these datasets, recent C-SLAM methods \cite{lajoie2023swarm, zhong2023dcl} continue to rely on splitting a single trajectory for validation.

In this paper, we introduce CU-Multi, a dataset specifically designed for evaluating methods in multi-robot data association, including collaborative SLAM and inter-robot loop closure detection. CU-Multi offers the following key features:
\begin{enumerate}
    \item \textit{A multi-robot dataset} consisting of two large-scale environments, each with four robots. Our dataset contains a total of eight diverse trajectories, spanning a combined length of 16.7 km across the CU Boulder campus (Figure~\ref{fig:main}).
    \item \textit{Systematically varied trajectory overlaps and a rendezvous-based trajectory design}, enabling evaluation of data association techniques under different levels of observational redundancy and field-relevant operational conditions.
    \item \textit{Semantic annotations} for all LiDAR scans, accurate geospatial pose alignment, and support code for easy interaction with the dataset.
\end{enumerate}

\section{RELATED WORK}
\begin{table*}
    \caption{Comparison of \textit{real-world} multi-robot Datasets. Note: only datasets with LiDAR have been listed. }
    \label{tab:multi_robot_datasets}
    \centering
    \begin{tabular}{>
    {\RaggedRight}m{2.2cm} >
    {\RaggedRight}m{1.5cm} >
    {\RaggedRight}m{2.8cm} >{\centering\arraybackslash}m{1.0cm} >{\centering\arraybackslash}m{0.6cm} >{\centering\arraybackslash}m{0.6cm} >
    {\centering\arraybackslash}m{1.0cm} >
    {\centering\arraybackslash}m{1.5cm} >{\centering\arraybackslash}m{1.5cm}} 
        \toprule
        \textbf{Dataset} & \textbf{\# Robots} & \textbf{Environments} & \textbf{RGB-D} & \textbf{GPS} & \textbf{IMU} & \textbf{Annotated LiDAR} & \textbf{Longest Trajectory (km)} & \textbf{Controlled Trajectory Overlap} \\ [0.5ex]
        \midrule 
        GRACO \cite{zhu2023graco} & 1 GW, 2 A & 8 Out & \cmark & \cmark & \cmark & \xmark & 0.82 & \xmark \\
        Kimera-Multi \cite{tian23arxiv_kimeramultiexperiments} & 8 GW & 1 Out, 1 In, 1 Hybrid & \cmark & \xmark & \cmark & \xmark & 1.40 & \xmark \\
        S3E  \cite{feng2024s3e} & 3 GW & 13 Out, 5 In & \cmark & \cmark & \cmark & \xmark & 1.95 & \cmark \\
        DiTer++ \cite{kim2024diter++} & 2 GL & 3 Out & \cmark & \xmark & \cmark & \xmark & N/A & \xmark \\
        Lamp 2.0 \cite{chang2022lamp} & 4 GW, 3 GL & 1 Out, 3 Sub & \cmark & \xmark & \cmark & \xmark & 2.20 & \xmark \\
        CoPeD \cite{zhou2024coped} & 3 GW, 2 A & 3 In, 3 Out & \cmark & \cmark & \cmark & \xmark & 0.36 & \xmark \\
        \textbf{CU-Multi (ours)} & \textbf{4 GW} & \textbf{2 Out} & \cmark & \cmark & \cmark & \cmark & \textbf{4.01} & \cmark \\[0.5ex]
        \bottomrule
    \end{tabular}
    \\[0.5ex]
    \footnotesize{\textbf{GW}, \textbf{GL}, and \textbf{A} denote ground-wheeled, ground-legged, and aerial  robots, respectively. \textbf{In}, \textbf{Out}, and \textbf{Hybrid} represent indoor, outdoor, and a mix of both, respectively.}
\end{table*}
While single-robot SLAM has long relied on curated datasets, multi-robot SLAM introduces challenges such as inter-robot data association that requires more representative data. Many works simulate multi-robot setups by splitting single-robot trajectories, but this approach raises concerns about both realism and consistency. This section reviews trajectory splitting strategies and existing multi-robot datasets, motivating the need for our proposed contribution.

\subsection{Trajectory Splitting on Single-Robot Datasets}
One of the most widely used datasets for evaluating single-robot SLAM and global localization pipelines is the KITTI Odometry Benchmark \cite{geiger2013vision}, a large-scale computer vision dataset collected in Karlsruhe, Germany using a sensor-equipped station wagon. It features stereo imagery, LiDAR, and GPS/IMU data across 22 driving sequences in urban and rural environments. In 2019, SemanticKITTI \cite{behley2019semantickitti} expanded the original KITTI dataset to include dense point-wise annotations for all lidar scans, enabling tasks such as semantic segmentation and scene completion. KITTI-360 \cite{liao2022kitti}, released in 2022, was also collected in Karlsruhe and contains broader city coverage, improved GPS accuracy, and consistent 2D/3D semantic instance annotations across panoramic imagery and LiDAR scans. Although originally intended for single-robot SLAM and perception, these datasets are commonly repurposed for verifying and benchmarking methods in multi-robot data association by artificially segmenting a single trajectory into multiple parts to simulate independent robot observations.

One of the earliest uses of trajectory splitting on modern SLAM datasets was introduced by Cieslewski and Scaramuzza in 2017 and 2018, who partitioned sequences from KITTI, the Málaga Urban Dataset \cite{blanco2014malaga}, and the MIT Stata Center Dataset \cite{fallon2013stata} into multiple segments to evaluate distributed visual localization and SLAM methods \cite{cieslewski2017efficient, cieslewski2018data}. Subsequent works adopted similar practices: Giamou et al. (2018) split KITTI sequences into two parts to evaluate communication-efficient loop closure \cite{giamou2018talk}, while DOOR-SLAM (2020) \cite{lajoie2020door}, DiSCO-SLAM (2021) \cite{huang2021disco}, and RDC-SLAM (2021) \cite{xie2021rdc} parsed KITTI sequences into 2–3 segments for evaluation.

In 2022, Lajoie et al. \cite{lajoie2022towards} cautioned that splitting trajectories at points of high observational overlap introduces unrealistic assumptions, as identical viewpoints and lighting conditions are unlikely to occur simultaneously in genuine multi-robot settings. Nevertheless, more recent methods continue this trend. DCL-SLAM (2023) \cite{zhong2023dcl} and Wi-Closure (2023) \cite{wang2023wi} both use partitioned KITTI sequences alongside limited field data. SWARM-SLAM (2024) \cite{lajoie2023swarm} splits both KITTI and KITTI-360 into 2–5 segments and supplements these with multi-robot datasets such as Graco \cite{zhu2023graco} and S3E \cite{feng2024s3e}.

Despite the widespread use of trajectory splitting, few works explicitly document their partitioning strategies. Moreover, no standardized partitioning protocol has been proposed or consistently referenced in the reviewed literature. Notably, FRAME \cite{stathoulopoulos2024frame}, a recent map-merging framework, split KITTI sequence \textit{00} into two sequential segments (0–1700 and 1701–4540). SideSLAM \cite{liu2024slideslam} evaluated their decentralized metric-semantic SLAM method by splitting SemanticKITTI sequences \textit{05} and \textit{07} into three overlapping parts: \textit{05}: (0–1000, 700–2000, 1600–2760) and \textit{07}: (0–600, 400–1000, 500–1100). Similarly, Cao et al. \cite{cao2024multi} validated their distributed multi-robot object SLAM approach by splitting KITTI sequences \textit{00}, \textit{05}, \textit{06}, and \textit{08} into three overlapping segments each: \textit{00}: (0–2000, 1500–3500, 2500–4540), \textit{05}: (0–1200, 800–2000, 1560–2760), \textit{06}: (0–700, 200–900, 400–1100), and \textit{08}: (0–2000, 1000–3000, 2000–4070). These methods not only partition individual trajectories in a way that could unrealistically lead to shared viewpoints and environmental conditions across the multi-robot trajectories they aim to represent, but the partitions can also overlap, including the \textit{exact same observations}. This continued reliance on single-robot benchmarks highlights a critical gap in evaluation practices: the failure to use standardized datasets that capture the observational diversity and independence inherent in multi-robot systems.

\subsection{Multi-Robot Datasets}
One of the earliest multi-robot datasets is the 2011 UTIAS dataset \cite{leung2011utias}, which includes nine indoor environments, each featuring five wheeled robots equipped with monocular cameras. In 2020, two additional datasets were introduced: FordAV \cite{agarwal2020ford} and AirMuseum \cite{dubois2020airmuseum}. FordAV provides multi-seasonal data collected by a fleet of Ford Fusion vehicles outfitted with four LiDARs, GPS, and six cameras. However, the vehicle trajectories exhibit near-complete overlap across all sequences, requiring users to manually segment them to simulate limited overlap scenarios. AirMuseum captures warehouse-scale data using a heterogeneous team of two ground robots and one aerial platform, each equipped with a stereo camera rig. While both AirMuseum and UTIAS offer multi-robot observations, their indoor collection settings and lack of LiDAR data limit their scalability and applicability to modern C-SLAM pipelines.

More recent datasets have broadened the scope of multi-robot data collection. GRACO \cite{zhu2023graco} introduced a heterogeneous two-robot team consisting of a ground and aerial robot; however, the dataset is constrained by a low-resolution 16-beam LiDAR and short trajectories. In follow-up work to Kimera-Multi \cite{tian2022kimera}, Kim et al. released a dataset comprising three sequences of eight ground robots operating in indoor, outdoor, and hybrid environments \cite{kim2023multi}. In 2024, the CoPeD dataset \cite{zhou2024coped} was introduced, addressing specific limitations of prior datasets in terms of sensor heterogeneity and environmental diversity. Kimera-Multi and CoPeD address valuable aspects needed for a multi-robot dataset, but are still limited to small trajectories and minimal observational overlap.

The need for datasets that explicitly explore varying degrees of shared observations has also gained attention. S3E \cite{feng2024s3e}, introduced in 2024, presents four distinct trajectory paradigms to study different levels of inter-robot trajectory overlap using a team of three robots. While this approach is a valuable step toward evaluating data association under varied trajectory overlap conditions, the trajectories remain limited in scale and contain relatively few intra-robot revisits or loop closures.

Despite recent progress in multi-robot datasets, many works have yet to adopt them for evaluation. Relying on idealized, single-trajectory conditions can obscure failure modes that emerge when robots operate independently, with varied trajectory overlap, differing viewpoints, and asynchronous data capture. As multi-robot SLAM continues to mature, there is a growing need for datasets that provide ground truth across multiple independently moving platforms to rigorously evaluate inter-robot data association, loop closure detection, and map merging strategies. Our dataset addresses this gap by offering multi-robot trajectories with explicitly controlled overlap levels and annotated LiDAR scans, enabling more realistic and robust evaluation of collaborative SLAM methods.


\begin{table*}
    \caption{Hardware Specifications.} 
    \label{hardware_specs}
    \centering
    \begin{tabular}{
    >{\centering\arraybackslash}m{3cm} 
    >{\centering\arraybackslash}m{4cm} 
    >{\centering\arraybackslash}m{3cm} 
    >{\centering\arraybackslash}m{2.0cm} 
    >{\centering\arraybackslash}m{1.5cm} 
    >{\centering\arraybackslash}m{1.5cm} 
    } 
        \toprule
        \textbf{Equipment} & \textbf{Model Name} & \textbf{Characteristics} & \textbf{Resolution} & \textbf{FoV} & \textbf{Sensor Rate}\\[0.5ex]
        \midrule
        LiDAR & Ouster OS1-64 & 200 m range & 64v × 1028h & 45° vertical & 20 Hz \\[0.5ex]
        \hline
        IMU & MicroStrain 3DM-GQ7-GNSS/INS & ±8 g & 300 dps & - & 400 Hz \\[0.5ex]
        \hline 
        Network Interface Modem & MicroStrain 3DM-RTK Modem & 2 cm,  0.1° accuracy  & - & - & 2 Hz\\[0.5ex]
        \hline
        GNSS Antennas & u-blox ANN-MB-00 & - & - & - & - \\[0.5ex] 
        \hline
        RGB-D & Intel Realsense D455 & RGB: Global Shutter & 1280 × 800 & 90° × 65° & 30 Hz\\[0.5ex]
        \hline
        Main Computer & Intel NUC & Intel i7 CPU @ 3.20GHz, 16GB RAM & - & - & - \\[0.5ex]
        \bottomrule
    \end{tabular}
\end{table*}

\section{The  CU-Multi Dataset}

\begin{table}[thpb]
    \centering
    \caption{Inter-Robot and Intra-Robot Trajectory Overlap}
    \label{tab:traj_overlap}
    \begin{tabular}{c|cc|cccc}
    \hline
    & & & \multicolumn{4}{c}{\textbf{Source Trajectory}} \\[0.5ex]
    \textbf{Env.} & & & \textbf{robot1} & \textbf{robot2} & \textbf{robot3} & \textbf{robot4} \\
    \hline
    \multirow{9}{*}{\rotatebox[origin=c]{90}{\textbf{Main Campus}}}
    & \multirow{9}{*}{\rotatebox[origin=c]{90}{\textbf{Target Trajectory}}} & & & & &\\
    & & \textbf{robot1} & \textbf{1136.41} &  788.43 &  538.11 &  246.89 \\
    & & & & & & \\
    & & \textbf{robot2} &  721.75 & \textbf{1373.37} &  618.37 &  377.98 \\
    & & & & & & \\
    & & \textbf{robot3} &  454.97 &  560.16 & \textbf{2792.06} & 1097.44 \\
    & & & & & & \\
    & & \textbf{robot4} &  293.19 &  352.95 & 1009.14 & \textbf{4005.75} \\
    & & & & & & \\
    \midrule
    \multirow{9}{*}{\rotatebox[origin=c]{90}{\textbf{Kittredge Loop}}} & \multirow{9}{*}{\rotatebox[origin=c]{90}{\textbf{Target Trajectory}}} & & & & &\\
    & & \textbf{robot1} & \textbf{1295.96} &  595.97 &  371.70 &  523.19 \\
    & & & & & & \\
    & & \textbf{robot2} &  676.53 & \textbf{1360.43} &  487.74 &  570.46 \\
    & & & & & & \\
    & & \textbf{robot3} &  344.49 &  497.79 & \textbf{1816.76} &  807.95 \\
    & & & & & & \\
    & & \textbf{robot4} &  517.04 &  610.40 &  704.56 & \textbf{2971.38} \\
    & & & & & & \\
    \hline
    \end{tabular}
    \footnotesize{Distance (meters) of non-time synchronized overlap along the source robot's trajectory with the target robot's trajectory. \textit{Note}: the diagonals represent a robot's complete trajectory. We consider trajectory overlap as the source robot's pose being within 1 meter radius of the target robot's pose.}
\end{table}

The CU-Multi dataset was collected using a single robotic platform over multiple sessions. This section describes the platform and sensors used for data collection, our novel approach to ensuring accurate ground-truth localization across multiple runs within each environment, our two-stage method for lidar annotation, the large-scale outdoor environments where data was gathered, and how the dataset is structured.

\subsection{Platform and Sensors}
\begin{figure}[thpb]
  \centering
  \includegraphics[width=1.0\linewidth]{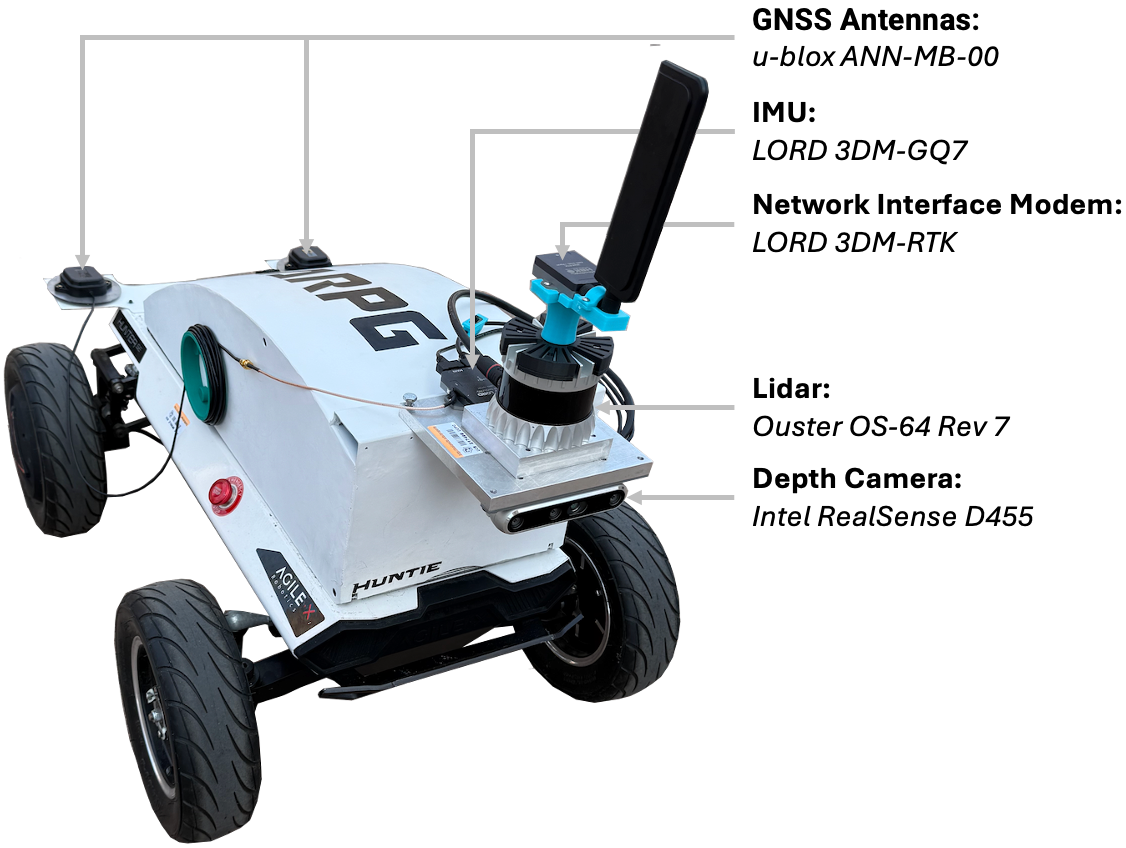}
  \caption{Hardware and sensor specifications used on platform.}
  \label{fig:sensors}
\end{figure}

The CU-Multi dataset was collected using the AgileX Hunter SE platform modified with a custom-designed electronics housing (see Figure~\ref{fig:sensors}). Inside the housing, the system includes an Intel NUC i7 with 16 GB of RAM, a power distribution board, an Ouster interface module, and a 217 Wh battery. Externally, a mounting plate on the rear of the Hunter SE accommodates two u-blox ANN-MB-00 GNSS antennas, while a front-mounted plate holds a 64-beam Ouster LiDAR sensor, a RealSense D455 RGB-D camera, a Lord MicroStrain G7 IMU, and a Lord RTK cellular modem with an external antenna.

\subsection{Dataset Collection and Environments}
The CU-Multi dataset was collected in two large outdoor environments in Boulder Colorado, at the University of Colorado Boulder (CU) campus. The first environment, \textit{main\_campus}, covers the central academic area, with approximately 7.4 km of traversed path. The second environment, \textit{kittredge\_loop}, encompasses a large region south of the main campus, offering more open space and varied terrain for multi-robot exploration.

Each of the two environments consists of four trajectories with varying levels of overlap, as seen in Figure \ref{fig:main}. We provide qualitative overlap metrics in Table~\ref{tab:traj_overlap}. In our dataset, \textit{robot 1} and \textit{robot2} share significant path overlap but capture the scene from different viewpoints, enabling users to test algorithms for viewpoint-invariant map merging and feature association. Meanwhile, \textit{robot3} and \textit{robot4} follow paths with less overlap, providing opportunities to evaluate performance under sparse or partially overlapping observations. The trajectory of \textit{robot3} extends the trajectory of the first two robots, introducing additional observations. Finally, \textit{robot4} encompasses all previous trajectories, covering the most extensive area. Formally, the relationship between these trajectories can be expressed as:
\begin{equation}
    T_1 \approx T_2, \quad (T_1 \cup T_2) \subseteq T_3, \quad (T_1 \cup T_2 \cup T_3) \subseteq T_4
 \end{equation}
where \( T_i \) represents the trajectory of \textit{robot\textbf{i}}.

\begin{figure}[thpb]
  \centering
  \includegraphics[width=1.0\linewidth]{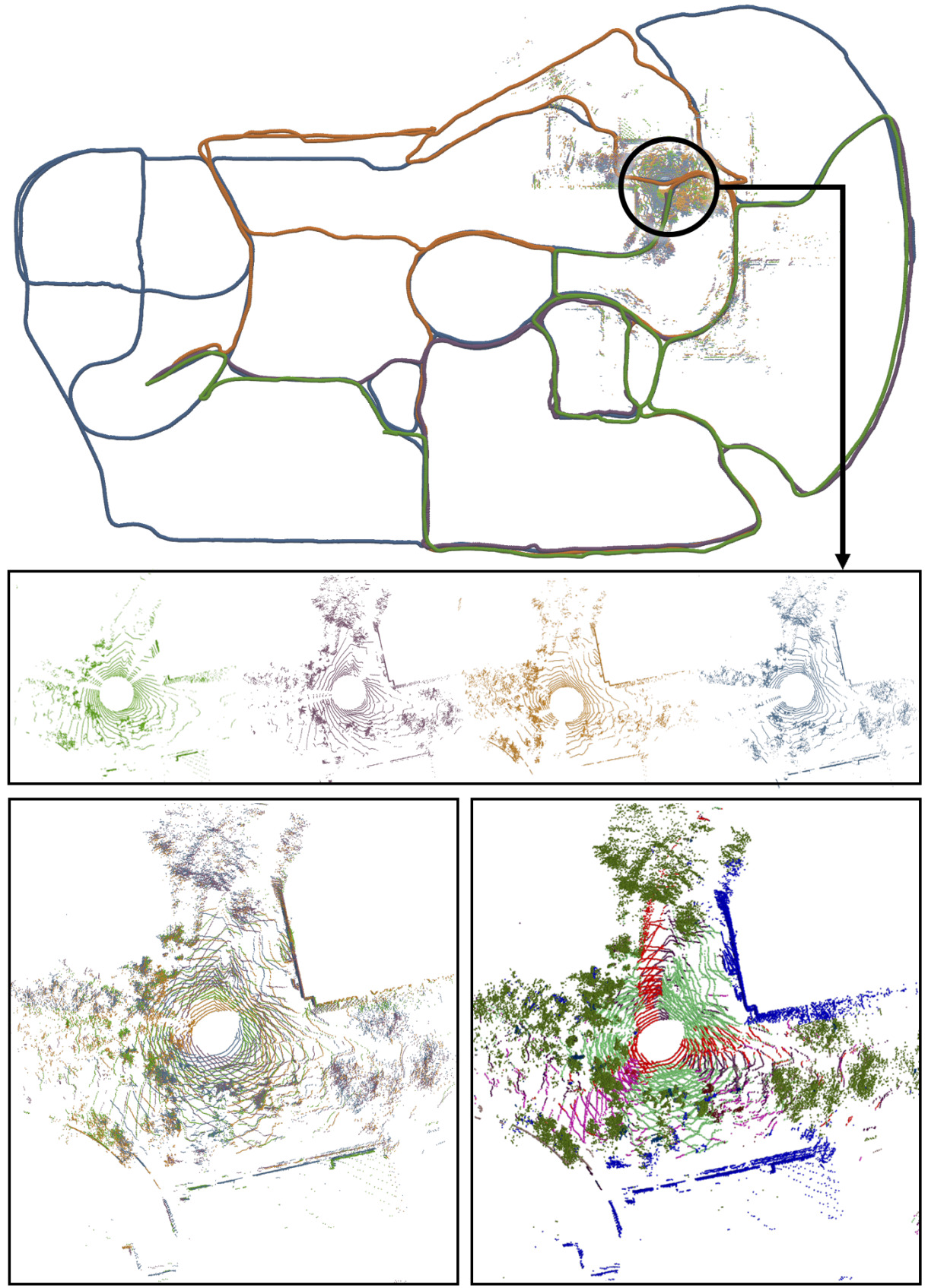}
  \caption{Visualization of the rendezvous point in the \textit{kittredge\_loop} environment. \textbf{Top:} All robot trajectories converge at a common location, enabling direct comparison of observations from different viewpoints. \textbf{Middle:} color-coded (key in Figure~\ref{fig:main}) raw LiDAR scans from each robot, highlighting how slight differences in robot pose can affect the distribution of points of the same physical scene. \textbf{Bottom-Left:} Overlapping and aligned color-coordinated scans at rendezvous point. \textbf{Bottom-Right:} Overlapping and aligned scans with semantic annotations.}
  \label{fig:rendezvous_example}
\end{figure}

To maximize viewpoint diversity, we introduced variations in observational perspectives while traversing overlapping regions as much as possible. The structured overlap across trajectories enables users to select data subsets based on their desired level of spatial redundancy and cross-view consistency. All trajectories end within distance of 4 meters of each other at a common location in each environment, representing a multi-robot rendezvous scenario (Figure~\ref{fig:rendezvous_example}).
\begin{figure}[thpb]
  \centering
  \includegraphics[width=1.0\linewidth]{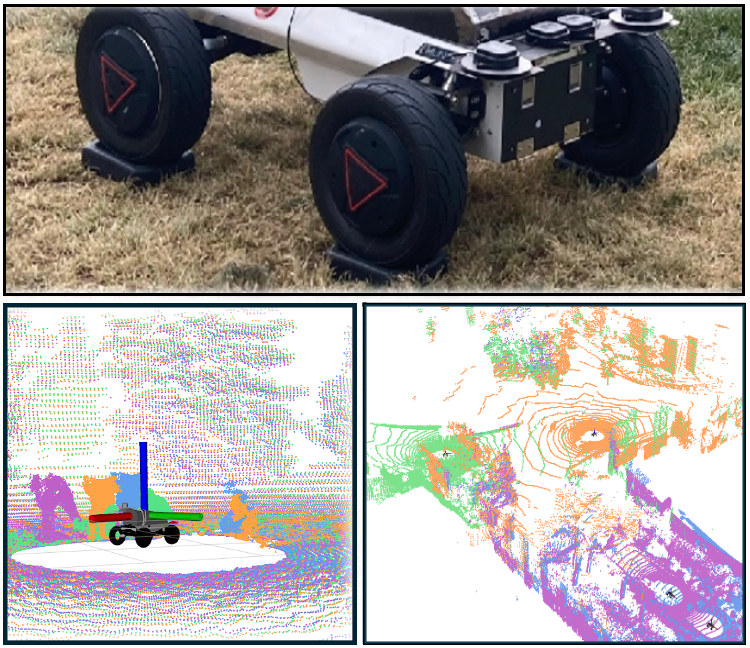}
  \caption{Our platform situated on wheel chocks before data collection (top). GPS aligned data playback for the entire quartet of robots  post-collection (bottom-left). Feature alignment between the quartet during data playback (bottom-right)}
  \label{fig:huntie_on_chocks}
\end{figure}

\subsection{Ground Truth Odometry and Geospatial Alignment}
\label{sec:gto_geospatial_alignment}
\begin{figure}[thpb]
  \centering
  \includegraphics[width=1.0\linewidth]{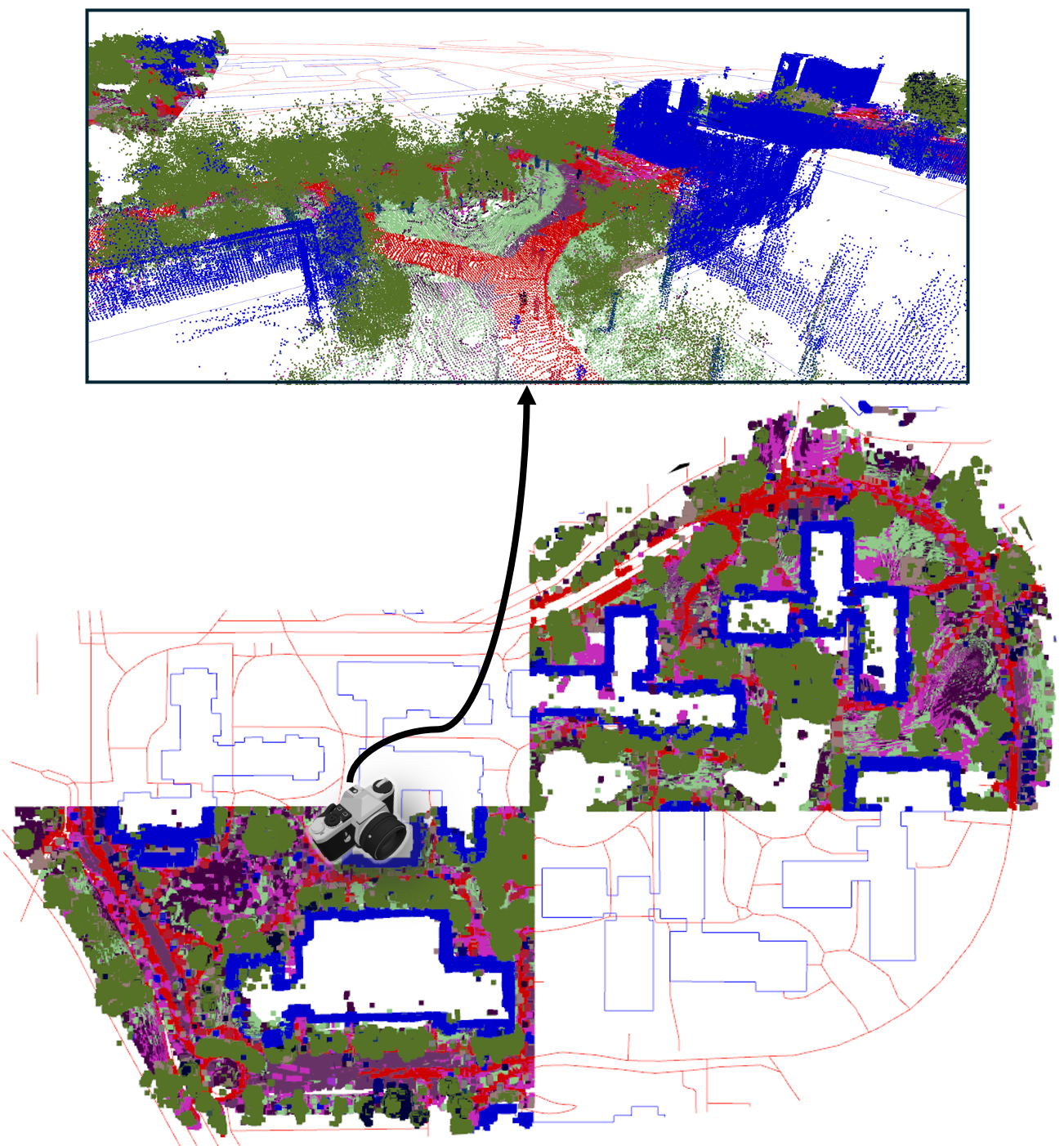}
  \caption{Example of geospatially-aligned poses through annotated lidar map overlaid onto OSM data (bottom) and example of 3D annotated lidar (top).}
  \label{fig:3d_semantics_aligned}
\end{figure}

To ensure precise odometry estimation, LiDAR timestamps were synchronized with the onboard IMU and computing system using Precision Time Protocol (PTP). We provide odometry for each run using a tightly-coupled configuration of LIO-SAM \cite{liosam2020shan} with RTK GPS and ensure accurate geospatial alignment across all collected data. Additionally, we include OpenStreetMap (OSM) \texttt{xml} files for each environment, along with sample code for performing projections (see Figure \ref{fig:3d_semantics_aligned}).

In addition to capturing centimeter-level accurate RTK measurements at 2Hz, we implemented an additional calibration procedure to further refine the relative starting positions of each run. Specifically, we initiated data collection from the same fixed point, ensuring consistency by securing four wheel chocks in the grass and positioning the four tires of the Hunter SE platform accordingly (Figure \ref{fig:huntie_on_chocks}). This provides an additional level of spatial consistency across runs, mitigating potential inaccuracies from initial GPS convergence delays and ensuring precise alignment of trajectory starting points for inter-run comparisons and ground truth alignment.

\subsection{LiDAR Annotation}
\label{sec:lidar_annotations}
None of the multi-robot datasets surveyed in this work provide ground truth semantic annotations for LiDAR data. Annotating large-scale LiDAR point clouds requires substantial time and resources, presenting a significant barrier to providing such data. To address this, we follow a similar approach as \cite{zhou2024coped}, which provides zero-shot annotations followed by a verification step. They infer 2D instance segmentation masks with SAM \cite{kirillov2023segment}, Grounding DINO \cite{liu2024grounding}, and RAM \cite{zhang2024recognize}, followed by temporal propagation \cite{yang2023track} to obtain consistent annotations over time.

In this work, we adopt a two-stage semantic labeling pipeline. In the first stage, we apply zero-shot inference to generate semantic labels for individual LiDAR scans. Specifically, we utilize CENet \cite{cheng2022cenet}, a LiDAR range image-based semantic segmentation network, to automatically annotate each scan according to the SemanticKITTI label taxonomy. Second, we leverage the tight geospatial alignment described in Section \ref{sec:gto_geospatial_alignment} to filter inferred semantics by ground truth labeled OSM.

\subsection{Dataset Structure}
\begin{figure}[thpb]
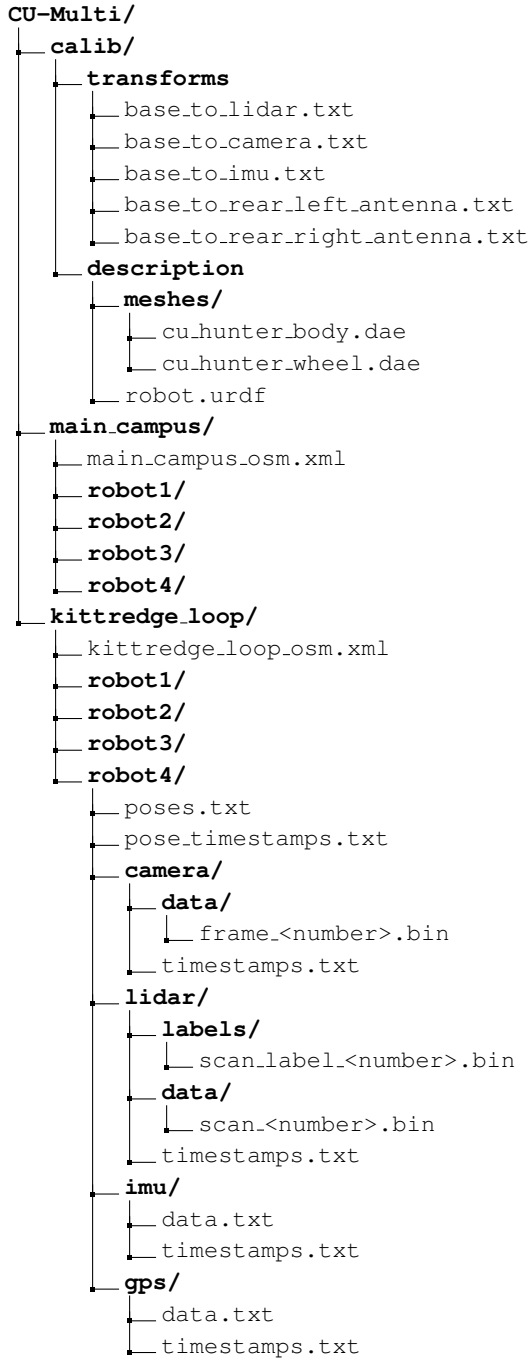

    \small
    \dirtree{%
    .1 \textbf{CU-Multi/}.
    .2 \textbf{calib/}.
    .3 \textbf{transforms}.
    .4 base\_to\_lidar.txt.
    .4 base\_to\_camera.txt.
    .4 base\_to\_imu.txt.
    .4 base\_to\_rear\_left\_antenna.txt.
    .4 base\_to\_rear\_right\_antenna.txt.
    .3 \textbf{description}.
    .4 \textbf{meshes/}.
    .5 cu\_hunter\_body.dae.
    .5 cu\_hunter\_wheel.dae.
    .4 robot.urdf.
    .2 \textbf{main\_campus/}.
    .3 main\_campus\_osm.xml.
    .3 \textbf{robot1/}.
    .3 \textbf{robot2/}.
    .3 \textbf{robot3/}.
    .3 \textbf{robot4/}.
    .2 \textbf{kittredge\_loop/}.
    .3 kittredge\_loop\_osm.xml.
    .3 \textbf{robot1/}.
    .3 \textbf{robot2/}.
    .3 \textbf{robot3/}.
    .3 \textbf{robot4/}.
    .4 poses.txt.
    .4 pose\_timestamps.txt.
    .4 \textbf{camera/}.
    .5 \textbf{data/}.
    .6 frame\_<number>.bin.
    .5 timestamps.txt.
    .4 \textbf{lidar/}.
    .5 \textbf{labels/}.
    .6 scan\_label\_<number>.bin.
    .5 \textbf{data/}.
    .6 scan\_<number>.bin.
    .5 timestamps.txt.
    .4 \textbf{imu/}.
    .5 data.txt.
    .5 timestamps.txt.
    .4 \textbf{gps/}.
    .5 data.txt.
    .5 timestamps.txt.
    }
    \caption{File structure for CU-Multi dataset.}
    \label{fig:dataset_structure}
\end{figure}

The dataset is structured hierarchically, with environments and a calibration directory at the top level. The calibration directory (\textit{calib}) contains the relative sensor positions on the platform, camera intrinsics, a mesh and URDF of the system. Within each environment, there are four directories corresponding to individual robots, labeled \textit{robot1} through \textit{robot4}. Each robot-specific directory follows a consistent structure with subdirectories containing data and timestamps for the onboard RGB-D camera, lidar, IMU, and GPS, as shown in Figure~\ref{fig:dataset_structure}. 

For users interested in testing algorithms through data playback, we provide Python scripts to convert the dataset into various formats, including \texttt{ROS} bags, \texttt{ROS 2} bags, and \texttt{MCAP} files. 

\section{PROPOSED DATASET USAGE}
The CU-Multi dataset is designed to facilitate research and development in multi-robot C-SLAM and map merging pipelines. Researchers can use this dataset to test and validate various algorithms related to loop closure detection, multi-robot localization, and map merging in real-world environments. Given its rich diversity of trajectories, the dataset offers a unique opportunity to evaluate the robustness of MRSs under different levels of path overlap and viewpoints. Additionally, the inclusion of zero-shot semantic segmentation labels and geospatial pose alignment makes it ideal for testing algorithms that require precise sensor data integration and semantics-based descriptor extraction. It is our hope that the CU-MULTI Dataset will be a valuable resource for evaluating C-SLAM techniques and exploring new approaches to robust inter-robot map merging.

\section{CONCLUSION AND FUTURE WORK}
In this work, we introduced CU-Multi, a multi-robot dataset collected using a single platform across two urban environments on the University of Colorado Boulder campus. The dataset is specifically designed to address core challenges in multi-robot C-SLAM, including inter-robot loop closure detection and map merging. CU-Multi features diverse trajectories, controlled levels of path overlap, and rich multi-modal sensor data, enabling robust evaluation of keyframe descriptors, data association methods, and localization strategies under realistic conditions. Looking ahead, there are several avenues for future work. First, we plan to extend CU-Multi by incorporating data collected simultaneously using multiple heterogeneous robotic platforms. This will enable validation of algorithms in scenarios involving diverse sensor configurations and dynamic inter-robot interactions. Second, we intend to provide illustrative baseline benchmarks using state-of-the-art multi-robot SLAM and map merging algorithms, offering the robotics community clear and standardized metrics for direct method comparison and performance evaluation.

\bibliographystyle{plain}
\bibliography{references}

\end{document}